%% file: index.tex
\documentclass[conference]{IEEEtran}
\usepackage{multirow}

\ifCLASSINFOpdf
\usepackage[pdftex]{graphicx}
\else
\fi

\usepackage{tikz}

\newcommand\copyrighttext{%
  \footnotesize  978-1-5090-0478-2/15/\$31.00 \copyright 2015 IEEE. Personal use of this material is permitted.
  Permission from IEEE must be obtained for all other uses, in any current or future
  media, including reprinting/republishing this material for advertising or promotional
  purposes, creating new collective works, for resale or redistribution to servers or
  lists, or reuse of any copyrighted component of this work in other works.
}

\newcommand\copyrightnotice{%
\begin{tikzpicture}[remember picture,overlay]
\node[anchor=south,yshift=10pt] at (current page.south) {\fbox{\parbox{\dimexpr\textwidth-\fboxsep-\fboxrule\relax}{\copyrighttext}}};
\end{tikzpicture}%
}

\begin{document}

%
\title{​Cosolver2B: An Efficient Local Search Heuristic for the Travelling Thief Problem}

\author{\IEEEauthorblockN{Mohamed El Yafrani}
\IEEEauthorblockA{LRIT, associated unit to CNRST (URAC 29)\\
Faculty of Science, Mohammed V University of Rabat\\
B.P. 1014 Rabat, Morocco\\
Email: m.elyafrani@gmail.com}
\and
\IEEEauthorblockN{Bela\"id Ahiod}
\IEEEauthorblockA{LRIT, associated unit to CNRST (URAC 29)\\
Faculty of Science, Mohammed V University of Rabat\\
B.P. 1014 Rabat, Morocco\\
Email: ahiod@fsr.ac.ma}}


%


\maketitle

\copyrightnotice

\begin{abstract}
Real-world problems are very difficult to optimize. 
However, many researchers have been solving benchmark problems 
that have been extensively investigated for the last decades 
even if they have very few direct applications. 
The Traveling Thief Problem (TTP) is a NP-hard optimization 
problem that aims to provide a more realistic model. 
TTP targets particularly routing problem under packing/loading 
constraints which can be found in supply chain management and 
transportation.
In this paper, TTP is presented and formulated mathematically. 
A combined local search algorithm is proposed and compared with 
Random Local Search (RLS) and Evolutionary Algorithm (EA).
The obtained results are quite promising since new better solutions were found.
\end{abstract}


%
\IEEEpeerreviewmaketitle

\input{sect1}

\input{sect2}
\input{sect3}
\input{sect4}

\input{sect5}


\section*{Acknowledgment}

The authors express their gratefulness to the following persons who have contributed 
to this work with their useful comments: Asmae El Ghazi, Aziz Ouaarab, and Mehdi El Krari.

\bibliographystyle{plain}
\bibliography{references}

\end{document}

%% file: sect1.tex
\section{Introduction}

The travelling thief problem (TTP) is a novel NP-hard problem introduced in 
\cite{ttp-2013} to provide a model that better represents real-world problems.
The particularity of TTP is that it is composed of two other NP-hard problems, 
namely the travelling salesman problem and the knapsack problem, which are interdependent.

The problem can be introduced with the following simplified statement:\\
\textit{"Given $n$ cities and $m$ items scattered among these cities, 
a thief with his rented knapsack should visit all $n$ cities, 
once and only once each, and pick up some items. 
The more the knapsack gets heavier, the more the thief becomes slower. 
What is the best path and picking plan to adopt to achieve the best benefit ?"}.

Two particular properties can be extracted from the statement above. 
First, the overall problem is composed of two sub-problems: 
choosing the best path and picking up the most profitable items. 
Second, the sub-problems are interdependent: when the knapsack gets heavier, 
the speed of the thief decreases. 
The composition increases the number of possible solutions, 
and the interdependence makes it impossible to isolate sub-problems.

This kind of multiple interdependent components is widely found in logistics 
and supply chain management. Nevertheless, few attempts have been made to 
study and solve these problems as a whole, and a lot of effort have been 
made to solve the components independently. 
In [3], the authors explain why there is a gap between the work done by the 
Evolutionary Computation researchers and real-world applications. 
As far as we know, these observations can be extended to all metaheuristics community.

Examples of other realistic problems with multiple interdependent sub-problems 
include most supply chain optimization problems \cite{single-silo-2012, multi-silo-2012},
routing problems with loading constraints such as 2L-CVRP and 3L-CVRP 
\cite{vr-load-2004,leung2013meta,bortfeldt2012hybrid}, 
and water tank delivery \cite{stolk2013combining}.

Since TTP was introduced, some algorithms were proposed to solve it. 
An Evolutionary Algorithm and a Random Local Search were proposed in \cite{ttp-benchmark-2014} 
to provide a starting point to other researchers. 
In \cite{socially-ttp-2014}, an approach named Cosolver was introduced to solve TTP by separating 
the sub-problems and managing a communication between them. 
Lastly, an interesting work on TTP introduces many complexity reduction 
techniques in order to solve very large instances in a time budget of 10 minutes 
\cite{mei2014improving}.

In this paper, we follow the ideas proposed in \cite{socially-ttp-2014} 
and \cite{mei2014improving} to propose 
further improvements and introduce an algorithm based on the Cosolver 
framework that can solve TTP instances efficiently.

This paper is organized as follows. In section \ref{sect2} TTP is mathematically 
defined and investigated. Section \ref{sect3} is dedicated to introduce our approach 
for solving TTP. The tests and results are presented in section \ref{sect4}. 
Section \ref{sect5} concludes the paper and outlines areas for future research.

%% file: sect2.tex
\section{Background}\label{sect2}

In this section, we present some background information about TTP. 
The problem is formulated and an abstract algorithm is presented.

\subsection{The Travelling Thief Problem}

Herein we formulate the Travelling Thief Problem (TTP) 
which combines two other well known benchmark problems, 
namely the Travelling Salesman Problem (TSP) and the Knapsack Problem (KP).

In TTP, we consider $n$ cities and the associated distance 
matrix $\{d_{ij}\}$. There are $m$ items scattered in these cities, 
each item $k$ have a profit $p_k$ and a weight $w_k$. 
A thief with his knapsack is going to visit all these cities (once and only once each), 
and pick up some items to fill his knapsack. 
We note $W$ the maximum capacity of the knapsack, $v_{min}$ and $v_{max}$ 
are the minimum and maximum possible velocity respectively. 

Also, we consider also the following constraints and parameters:

\begin{itemize}
\item Each item is available in only one city. We note $\{A_i\}$ the availability vector, 
$A_i \in \{1,..., n\}$ contains the reference to the city that contains the item $i$.
\item We suppose that the knapsack is rented. We note $R$ the renting price per time unit.
\item The velocity of the thief changes accordingly to the knapsack weight. 
We note $v_{x_i}$ the velocity of the thief at city $x_i$ (see equation \ref{eq:velocity}).
\begin{equation} \label{eq:velocity}
v_{x_i} = v_{max} - C*w_{x_i}
\end{equation}
where $C = \frac{v_{max}-v_{min}}{W}$ is a constant value, and $w_{x_i}$ represents the weight of the knapsack 
at city ${x_i}$.
\end{itemize}

The goal of the problem is to find the tour $x$ and the picking plan $z$ 
that optimize the total travel gain defined in equation \ref{eq:objective}.
\begin{equation} \label{eq:objective}
G(x, z) = g(z) - R*f(x, z)
\end{equation}
Where $g(z) = \sum_m p_m * z_m$ is the total value of the items subject to $\sum_m w_m * z_m \le W$,
and $f(x, z) = \sum_{i=1}^{n-1} t_{x_i, x_{i+1}} + t_{x_n, x_1}$ is the total travel time.

The solutions could be naturally coded as follows. 
The tour $x = (x_1, ..., x_n)$ is a vector containing the ordered list of cities, 
and the picking plan $z = (z_1, ..., z_n)$ is a binary vector such as $z_i$ is 
equal to $1$ if the item $i$ is picked, $0$ otherwise.

The interdependence between KP and TSP have been investigated. 
As shown in \cite{ttp-2013,mei2014improving}, optimizing the sub-problems 
in isolation (even to optimality) does not guarantee finding good solutions 
for the overall problem. Therefore, finding good global solutions requires 
an algorithm that takes interdependence of components in consideration, 
which makes the design of such an algorithm quite difficult.

\subsection{Cosolver}

\begin{figure*}[ht]\centering
\includegraphics[width=.9\linewidth]{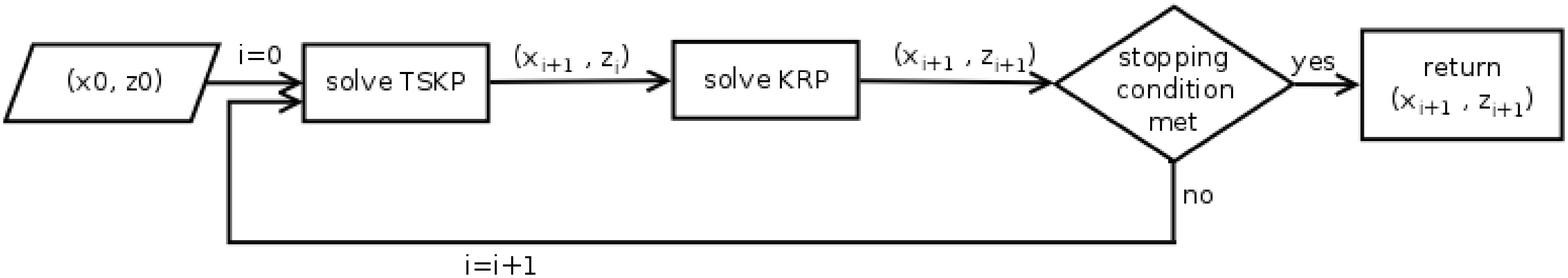}
\caption{A simplified Cosolver flowchart}
\label{fig:cosolver}
\end{figure*}

Cosolver is an abstract algorithm proposed in \cite{socially-ttp-2014} to solve TTP. 
The idea behind the algorithm is simple. Decompose the overall problem and solve 
components separately using a fitness function that takes in consideration the interdependence 
between the two components. 

Thus, given an initial solution $(x_0, z_0)$, TTP is decomposed into the following problems.

\subsubsection{The Travelling Salesman with Knapsack Problem (TSKP)}
consists of finding the best tour $x'$ that, combined with the last found picking plan $z'$, 
optimizes the TTP objective function $G$. Also, because the total profit function $g$ 
does not depend on the tour, instead of maximizing $G$, 
we can consider minimizing the total travel cost $T$ (see equation \ref{eq:tskp-fitness}).
\begin{equation} \label{eq:tskp-fitness}
T(x, z) = R*f(x, z)
\end{equation}

\subsubsection{The Knapsack on the Route Problem (KRP)}
consists of finding the best picking plan $z'$ that maximizes the total gain $G$ 
when combined with the last found tour $x'$.

Figure \ref{fig:cosolver} represents a flowchart of the Cosolver framework.

%% file: sect3.tex
\section{Proposed approach}\label{sect3}

In this section, we propose an algorithm that implements the Cosolver framework 
and uses some techniques proposed in \cite{mei2014improving}.

\subsection{Solution initialization}

In TTP, the initial solution has a big impact on the search algorithm. 
Thus, the initialization strategy should be chosen carefully. 
In our implementation, we use a heuristic approach to initialize both the 
tour and the picking plan.

Firstly, the tour is generated using a good TSP algorithm such as the Lin-Kernighan 
heuristic \cite{helsgaun2000effective} or the Ant Colony Optimization algorithm \cite{dorigo1997ant}.

Then, the picking plan is initialized using the following approach.
\begin{enumerate}
\item The insertion heuristic proposed in \cite{mei2014improving} is used to fill 
the knapsack with items according to three fitness approximations.
\item A simple bit-flip search on inserted items is performed to eliminate some useless 
items from the picking plan using the objective function.
\end{enumerate}

We will refer to this heuristic as the \textit{insertion \& elimination heuristic}.

\subsection{Complexity reduction techniques}

Our algorithm uses the following complexity reduction techniques.

\subsubsection{TSKP neighborhood reduction}

The Delaunay triangulation \cite{delaunay1934sphere} is used as a candidate generator 
for the 2-OPT heuristic. This strategy was also proposed in \cite{mei2014improving}.

\subsubsection{Objective value recovery}

Instead of using the objective function to calculate the objective value of neighbors, 
this value can be recovered by keeping track of time and weight information at each 
tour's city. The following vectors are used to perform such operation.

\begin{itemize}
\item Time accumulator ($t^{acc}$): a vector that contains the current time at each city of the tour.
\item Weight accumulator ($w^{acc}$): a vector that contains the current weight at each city of the tour.
\item Time register ($t^{reg}$): a vector that contains the added time at each city of the tour.
\item Weight register ($w^{reg}$): a vector that contains the added weight at each city of the tour.
\end{itemize}

\begin{figure}[ht]\centering
\includegraphics[width=.95\linewidth]{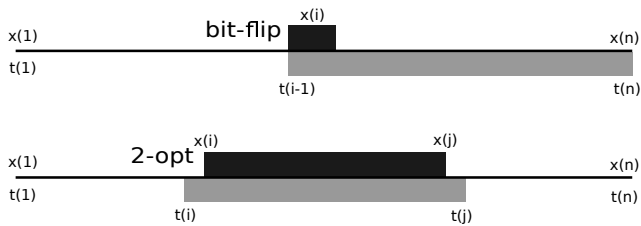}
\caption{Example of changes that can be applied to a TTP solution (in black) and its effect on the total traveling time function (in gray)}
\label{fig:ttp-change}
\end{figure}

Figure \ref{fig:ttp-change} shows the effect (in gray) of mutating a TTP solution on the 
total travel time. Two mutations are shown (in black): 
the one-bit-flip on the picking plan, and a 2-OPT exchange on the tour.

A similar technique named incremental evaluation was proposed in \cite{mei2014improving}. 
In our implementation, we go further and use these techniques also to recover the vectors.

\subsubsection{Objective function calculation}

The objective function has a complexity of $O(m*n)$ as proposed in \cite{ttp-investigation-2014}, 
where $m$ is the number of items and $n$ is the number of cities. 
The complexity could be reduced to $O(k*n)$ where $k$ is the number of items per city. 
In the CEC '2015 competition website \footnote{http://cs.adelaide.edu.au/~optlog/CEC2015Comp/}, 
the Java implementations already has a reduced complexity but it uses a technique highly 
dependent on the TTP instance generator. 
We believe that a better way is to class items per city. 

\begin{figure}[ht]\centering
\includegraphics[width=.95\linewidth]{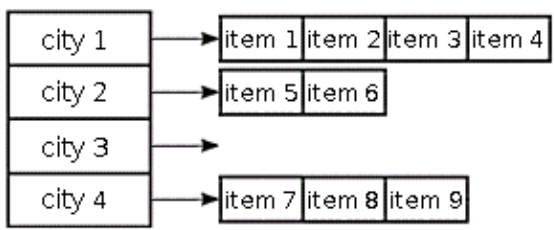}
\caption{Data structure for classing items by cities}
\label{fig:classed-items}
\end{figure}

Figure \ref{fig:classed-items} explains this technique using a simple example.
In the example, 4 cities are considered: \textit{city 1} contains 4 items (1, 2, 3 and 4),
\textit{city 2} contains 2 items (5 and 6), \textit{city 3} contains no items,
\textit{city 4} contains 3 items (7, 8 and 9).

\subsection{Proposed algorithm}

\begin{figure}[ht]\centering
\includegraphics[width=.6\linewidth]{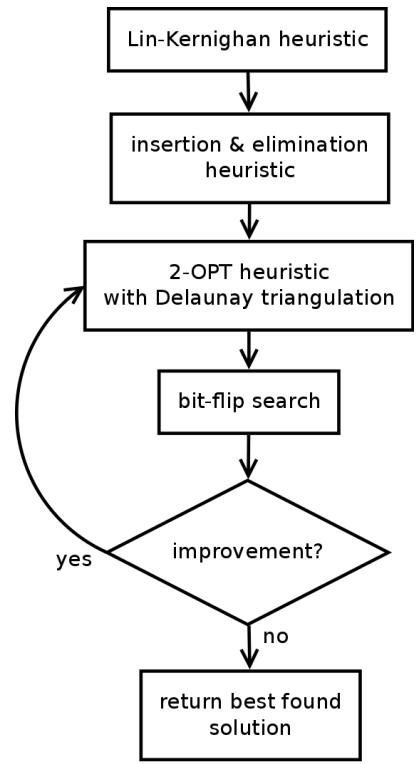}
\caption{Cosolver2B: Cosolver with 2-opt and bit-flip}
\label{fig:cosolver2b-flowchart}
\end{figure}

Our heuristic, namely Cosolver2B, is based on local search algorithms, 
and it implements the Cosolver framework. In addition to the techniques 
presented above, we use the 2-OPT heuristic to solve TSKP 
and a bit-flip search to solve the KRP.
The algorithm is summarized in the flowchart in figure \ref{fig:cosolver2b-flowchart}.

The algorithm starts with a tour generated using the Lin-Kernighan heuristic.
The insertion \& elimination heuristic is then used to initialize the picking plan.
The obtained tour and picking plan are then given to the TSKP optimizer
that tries to improve the current tour using a 2-OPT heuristic. 
The resulting tour is combined with the current picking plan and given to the KRP optimizer
that uses a simple bit-flip to find a decent picking plan. If there is improvement, 
the new picking plan is combined with the current tour and given to the TSKP optimizer.
This process is repeated until there is no improvement.

%% file: sect4.tex
\section{Experimental results}\label{sect4}

\begin{table*}
  \normalsize
  \caption{Results of Cosolver2B compared to EA and RLS\label{tab:results}}
  \center
  \begin{tabular}{|l|llll|llll|}
    \hline
    \multirow{2}{*}{\textbf{instance}}      & \multicolumn{2}{l}{\textbf{EA}}   & \multicolumn{2}{l|}{\textbf{RLS}}  & \multicolumn{2}{l}{\textbf{Cosolver2B} \scriptsize{firstfit}}  & \multicolumn{2}{l|}{\textbf{Cosolver2B} \scriptsize{bestfit}} \\
                           & \footnotesize{objective} & \footnotesize{time} & \footnotesize{objective} & \footnotesize{time} & \footnotesize{objective} & \footnotesize{time} & \footnotesize{objective} & \footnotesize{time} \\ \hline
    \footnotesize{berlin52\_n255\_uncorr-similar-weights\_05}    & 20591    & 1.28   & 20591    & 0.89       & 25440     & 0.13  & \textbf{26658} & 0.11 \\ \hline
    \footnotesize{kroA100\_n495\_uncorr-similar-weights\_05}     & 38864    & 1.92   & 38864    & 1.06       & \textbf{39599}     & 0.21  & 39597 & 0.27 \\ \hline
    \footnotesize{ch150\_n745\_uncorr-similar-weights\_05}       & 40922    & 2.69   & 40922    & 1.38       & 52225     & 0.38  & \textbf{53075} & 0.43 \\ \hline
    \footnotesize{ts225\_n1120\_uncorr-similar-weights\_05}      & 84907    & 3.82   & 84907    & 1.69       & 88925     & 0.61  & \textbf{89079} & 0.92 \\ \hline
    \footnotesize{a280\_n1395\_uncorr-similar-weights\_05.ttp}   & 104365   & 5.64   & 104365   & 1.84       & 107233 &  0.88    & \textbf{112003} & 1.09 \\ \hline
    \footnotesize{lin318\_n1585\_uncorr-similar-weights\_05}     & 75387    & 6.87   & 75387    & 3.81       & \textbf{114915}    & 1.25  & \textbf{114915} & 1.54 \\ \hline
    \footnotesize{u574\_n2865\_uncorr-similar-weights\_05}       & 223165   & 21     & 223165   & 5.42       & 243571    & 5.36  & \textbf{253770} & 6.51 \\ \hline
    \footnotesize{dsj1000\_n4995\_uncorr-similar-weights\_05}    & 320165   & 47     & 320165   & 12         & 332916    & 5.95  & \textbf{332917} & 8.44 \\ \hline
    \footnotesize{rl1304\_n6515\_uncorr-similar-weights\_05}     & 573667   & 95     & 573667   & 21         & 571634    & 68    & \textbf{586576} & 83 \\ \hline
    \footnotesize{fl1577\_n7880\_uncorr-similar-weights\_05}     & 589756   & 127    & 589756   & 28         & 665866    & 139   & \textbf{684731} & 148 \\ \hline
    \footnotesize{d2103\_n10510\_uncorr-similar-weights\_05}     & 801110   & 214    & 801110   & 44         & \textbf{869584}    & 357   & \textbf{869584} & 389 \\ \hline
    \footnotesize{pcb3038\_n15185\_uncorr-similar-weights\_05}   & 1119168  & 427    & \textbf{1119169}  & 101        & 1117659   & 600   & 1086092 & 600 \\ \hline
    \footnotesize{fnl4461\_n22300\_uncorr-similar-weights\_05}   & 1477673  & 600    & \textbf{1478963}  & 265        & 1220469   & 600   & 696650  & 600 \\ \hline
    \footnotesize{rl11849\_n59240\_uncorr-similar-weights\_05}   & 2152954  & 600    & \textbf{4328939}  & 600        & 1785671   & 600   & 478884  & 600 \\ \hline
    \footnotesize{usa13509\_n67540\_uncorr-similar-weights\_05}  & 2226676  & 600    & \textbf{5077892}  & 600        & 4420300   & 600   & 4082169  & 600 \\ \hline
    \footnotesize{pla33810\_n169045\_uncorr-similar-weights\_05} & -9929644 & 600    & -3158915 & 600        & 14887314  & 600   & \textbf{15085752} & 600 \\ \hline
  \end{tabular}
\end{table*}

We tested our algorithm on 15 TTP instances of various sizes \cite{ttp-benchmark-2014}. 
The obtained results are compared with RLS and EA proposed in the same paper 
\footnote{for instance files and raw results of RLS and EA, 
the reader is referred to https://sites.google.com/site/mohammadrezabonyadi/standarddatabases/travelling-thief-problem-data-bases-and-raw-results}.

We have implemented two versions of Cosolver2B. 
The first is \textit{best fit} (also known as best improvement) which searches the 
entire neighborhood on a given iteration and selects the best neighbor. 
The second is \textit{first fit} (also known as first improvement) which stops the 
neighborhood search once a better solution is found.

Since our algorithms are deterministic, one run per instance is sufficient. 
The algorithms have a maximum runtime limit of $600$ seconds. 
Note that our implementation uses the Java platform, and the tests are performed on a 
core i3-2370M CPU 2.40GHz machine with $4$ GB of RAM, running Linux.

On the other hand, due to their random behavior, $30$ runs per instance were performed 
using RLS and EA. The tests on EA and RLS were performed on a different machine 
(Matlab 2014, core i7 2600 CPU 3.4 GHz, with 4GB of RAM, running Windows 7). 
Thus, The runtimes for these algorithms are given for guidance. 
Furthermore, in our comparison, we only select the best found solution for these two algorithms.

The results are reported in table \ref{tab:results}. 
Note that the runtimes are measured with seconds and the best objective 
values are made bolder.

The results show that our algorithm was able to find new better solutions for most tested 
instances only by combining local search algorithms. 
Furthermore, we have made the following observations:

\begin{itemize}
\item Cosolver2B surpasses EA and RLS on various TTP sizes.
\item The runtime of Cosolver2B is very decent for small and mid-size instances. However, for very large instances the runtime is quite high compared to RLS and EA.
\item Both EA and RLS have an unpredictable behavior, while Cosolver2B is deterministic and garantees decent solutions for most instances.
\item For many instances, most optimization runtime and gain improvement is done on the KRP sub-problem.
\item The best fit strategy performed better than first fit for most small and mid-size instances.
\end{itemize}

%% file: sect5.tex
\section{Conclusion}\label{sect5}

In this paper, a combined local search heuristic has been proposed. 
Our approach is based on the Cosolver framework and complexity reduction techniques. 
The experimental results have shown promising results both in terms of 
solution quality and runtime. 
The good performance of Cosolver2B and the memetic algorithm MALTS 
on very large scale instances \cite{mei2014improving}
proves the importance of making strong assumptions about the problem domain
when dealing with multi-component problems. 

Furthermore, we engage to improve our algorithm in terms of space exploration 
and runtime by using more sophisticated search heuristics.

%% file: index.bbl
\begin{thebibliography}{10}

\bibitem{ttp-2013}
Mohammad~Reza Bonyadi, Zbigniew Michalewicz, and Luigi Barone.
\newblock The travelling thief problem: the first step in the transition from
  theoretical problems to realistic problems.
\newblock In {\em Evolutionary Computation (CEC), 2013 IEEE Congress on}, pages
  1037--1044. IEEE, 2013.

\bibitem{socially-ttp-2014}
Mohammad~Reza Bonyadi, Zbigniew Michalewicz, Micha{\u{o}}
  Roman~Przyby{\u{o}}ek, and Adam Wierzbicki.
\newblock Socially inspired algorithms for the travelling thief problem.
\newblock In {\em Proceedings of the 2014 conference on Genetic and
  evolutionary computation}, pages 421--428. ACM, 2014.

\bibitem{bortfeldt2012hybrid}
Andreas Bortfeldt.
\newblock A hybrid algorithm for the capacitated vehicle routing problem with
  three-dimensional loading constraints.
\newblock {\em Computers \& Operations Research}, 39(9):2248--2257, 2012.

\bibitem{delaunay1934sphere}
Boris Delaunay.
\newblock Sur la sphere vide.
\newblock {\em Izv. Akad. Nauk SSSR, Otdelenie Matematicheskii i Estestvennyka
  Nauk}, 7(793-800):1--2, 1934.

\bibitem{dorigo1997ant}
Marco Dorigo and Luca~Maria Gambardella.
\newblock Ant colonies for the travelling salesman problem.
\newblock {\em BioSystems}, 43(2):73--81, 1997.

\bibitem{helsgaun2000effective}
Keld Helsgaun.
\newblock An effective implementation of the lin--kernighan traveling salesman
  heuristic.
\newblock {\em European Journal of Operational Research}, 126(1):106--130,
  2000.

\bibitem{single-silo-2012}
Maksud Ibrahimov, Arvind Mohais, Sven Schellenberg, and Zbigniew Michalewicz.
\newblock Evolutionary approaches for supply chain optimisation: part i: single
  and two-component supply chains.
\newblock {\em International Journal of Intelligent Computing and Cybernetics},
  5(4):444--472, 2012.

\bibitem{multi-silo-2012}
Maksud Ibrahimov, Arvind Mohais, Sven Schellenberg, and Zbigniew Michalewicz.
\newblock Evolutionary approaches for supply chain optimisation. part ii:
  multi-silo supply chains.
\newblock {\em International Journal of Intelligent Computing and Cybernetics},
  5(4):473--499, 2012.

\bibitem{vr-load-2004}
Manuel Iori and Silvano Martello.
\newblock Routing problems with loading constraints.
\newblock {\em Top}, 18(1):4--27, 2010.

\bibitem{leung2013meta}
Stephen~CH Leung, Zhenzhen Zhang, Defu Zhang, Xian Hua, and Ming~K Lim.
\newblock A meta-heuristic algorithm for heterogeneous fleet vehicle routing
  problems with two-dimensional loading constraints.
\newblock {\em European Journal of Operational Research}, 225(2):199--210,
  2013.

\bibitem{mei2014improving}
Yi~Mei, Xiaodong Li, and Xin Yao.
\newblock Improving efficiency of heuristics for the large scale traveling
  thief problem.
\newblock In {\em Simulated Evolution and Learning}, pages 631--643. Springer,
  2014.

\bibitem{ttp-investigation-2014}
Yi~Mei, Xiaodong Li, and Xin Yao.
\newblock On investigation of interdependence between sub-problems of the
  travelling thief problem.
\newblock {\em Soft Computing}, pages 1--16, 2014.

\bibitem{ttp-benchmark-2014}
Sergey Polyakovskiy, Mohammad Reza, Markus Wagner, Zbigniew Michalewicz, and
  Frank Neumann.
\newblock A comprehensive benchmark set and heuristics for the traveling thief
  problem.
\newblock In {\em Proceedings of the Genetic and Evolutionary Computation
  Conference (GECCO), Vancouver, Canada}, 2014.

\bibitem{stolk2013combining}
Jacob Stolk, Isaac Mann, Arvind Mohais, and Zbigniew Michalewicz.
\newblock Combining vehicle routing and packing for optimal delivery schedules
  of water tanks.
\newblock {\em OR Insight}, 26(3):167--190, 2013.

\end{thebibliography}
